\documentclass[sigconf,screen]{acmart}
\setcopyright{none}
\renewcommand\footnotetextcopyrightpermission[1]{}
\usepackage{xcolor}
\usepackage{url}

\usepackage{amsmath}
\usepackage{booktabs}
\usepackage{tabularx}
\usepackage{color}
\usepackage{graphics}
\usepackage{graphicx}
\usepackage{enumitem}
\usepackage{balance}

\usepackage{hyperref}
\hypersetup{
    colorlinks=true,
    linkcolor=blue,
    filecolor=magenta,
    urlcolor=cyan,
    breaklinks=true
}
\usepackage{cleveref}

\definecolor{amber}{rgb}{1.0, 0.75, 0.0}
\newcommand{\topic}[1]{}

\newtheorem{challenge}{Research Challenge}

\title{Pitfalls of Explainable ML: An Industry Perspective}

\author{Sahil Verma}
\authornote{These authors contributed equally to the paper.\vspace{-0.5em}}
 \affiliation{University of Washington, Arthur AI}
 \email{vsahil@cs.washington.edu}

\author{Aditya Lahiri}
\authornotemark[1]
 \affiliation{BITS Pilani, Goa Campus}
 \email{adityalahiri13@gmail.com}

 \author{John P.\ Dickerson}
 \affiliation{University of Maryland, Arthur AI}
 \email{john@cs.umd.edu}
 
 \author{Su-In Lee}
 \affiliation{University of Washington}
 \email{suinlee@cs.washington.edu}

\begin{document}

\begin{abstract}
As machine learning (ML) systems take a more prominent and central role in contributing to life-impacting decisions, ensuring their trustworthiness and accountability is of utmost importance. 
Explanations sit at the core of these desirable attributes of a ML system. The emerging field is frequently called ``Explainable AI (XAI)'' or ``Explainable ML.'' 
The goal of explainable ML is to intuitively explain the predictions of a ML system, while adhering to the needs to various stakeholders. 
Many explanation techniques were developed with contributions from both academia and industry. 
However, there are several existing challenges that have not garnered enough interest and serve as roadblocks to widespread adoption of explainable ML. 
In this short paper, we enumerate challenges in explainable ML from an industry perspective.
We hope these challenges will serve as promising future research directions, and would contribute to democratizing explainable ML.


\end{abstract}

\maketitle

\section{Introduction}
ML models can be broadly categorized into inherently explainable (e.g. linear models, decision trees) and black-box (e.g. random forest, neural network) models. 
The predictions of black-box models may require \emph{explanation}, as the models themselves are too complex to comprehend. 
There exist various explainable ML techniques for such models, generating explanations for the different data modalities (tabular, image, and textual data). 
Explanations are generated either at global or local level. 
Global explanations entail the explanation of the overall model, whereas the local one generates explanations for a single data-point. 
Explainable ML is an active area of research in academia and industry alike, and has desirable applications in the entire ML model cycle, i.e. during training, validation, deployment, monitoring, and updation. 
Explainable ML is useful for many stakeholders, like developers of the model, program managers, users of the tool. 
Although many popular explainable ML methods are used in practice, the problem is far from solved. 
In this short paper, we enumerate the current hindering challenges the usage of explainable ML in an industrial setting. 
\emph{Our team has worked variously at startups, public companies, non-profits, in fintech and healthcare. We put this short paper accumulating from our experience of using explainability at such places.}

\begin{challenge}\label{ch:interpret}
Inconsistent language of explainable ML techniques. 
\end{challenge}

Explainable ML techniques do not share a common vocabulary. 
Even for a single data modality like tabular data, explanations are produced in myriad formats like rule sets, feature attributions, and heat-maps. 
The decision of which explanation is to be used is left to the end-user, which in several situations are non-technical users. 

The inconsistency in formats makes it difficult for end-users who have to ingest these explanations for downstream tasks and make decisions based on them.
This raises questions about the actual contribution of explainable ML~\cite{alufaisan2020does}, while it might be an artifact of the current research in explainable ML and not the field itself. 
Future research in developing a shared language for explainable ML, at least for each data modality would help the end-users. 
This language should be developed while keeping all stakeholders in mind, specially the non-technical audience. 


\begin{challenge}\label{ch:evaluating}
Lack of quantitative evaluation metrics. 
\end{challenge}
Different explainable ML techniques provide explanations in different ways with few quantitative metrics to compare them. This is specially challenging if two techniques are a minor variant of each other. 
For example, LIME~\cite{lime} is a explainablity technique which provides explainability in form of feature attributions. 
xLIME~\cite{xlime} is a variant of LIME, which has a slightly different strategy to generate explanations. 
If an end-user is presented with explanations from both LIME and xLIME, there exists no method which can quantitatively state which of the two are more useful in different use cases (although some research exists to measure which of them is more faithful~\cite{Gagan-staining}).  
The same is true for SHAP~\cite{shap} and its variants~\cite{shap-variant}. 

\citet{doshivelez2017rigorous} discuss the taxonomy of interpretability evaluation in three ways. 
Two of these involve a human, who has to judge the usefulness of the explanations in a downstream task. Conducting human experiments might not be feasible in all scenarios due to lack of resources or expertise. 
In the third way, no human in involved. A proxy task is decided and human behavior is simulated. 
However, none of these are quantitative metrics for comparison across techniques. 

Currently, industry uses metrics like fidelity of the surrogate model to the original model, robustness of the explanation~\cite{bhatt-aggregating,nguyen2020quantitative}, and precision of rules~\cite{anchors} as quantifiable metrics of performance of a technique~\cite{choose-explainer}. 
But these might not reflect the actual usefulness of the explanations~\cite{papenmeier2019model}. 
Some works have proposed a few quantifiable metrics for comparing explainability techniques~\cite{aggregation-tree,nguyen2020quantitative}, however an agreement on a single representative metric does not exist. 

An emerging explainability technique, counterfactual explanations come with several quantifiable metrics~\cite{verma2020counterfactual} to judge the usefulness of an explanation. These metrics have become great choices for algorithms to optimise for. 




\begin{challenge}\label{ch:adapt}
Lack of established workflow for several use cases. 
\end{challenge}
As mentioned earlier, explainable ML is of use in all stages of a ML model cycle~\cite{chi-xai2020}. 
Yet, most explainability techniques are only developed to understand the prediction of a given model on a single datapoint or a collection of datapoints. 
If a developer wants to debug or monitor a model, there do not exist established guidelines which they can follow~\cite{klaise2020monitoring} to be helped by explainable ML techniques. 
The evaluation of explainability techniques also needs to consider this real-world usage of explainability, contrary to the current metrics like fidelity. 


\begin{challenge}\label{ch:scalability}
Lack of scalable implementations. 
\end{challenge}

The open-source implementation of widely used explainable ML frameworks like LIME and SHAP are not efficient enough for big data. 
\citet{shap-complexity} have discussed the NP-hardness of SHAP explainability for models like logistic regression and neural networks with sigmoid activation. 
Most explainability implementations are not built to explain models built using distributed systems like PySpark~\cite{pyspark-Shap}. 
This severely restricts their usage in organizations which have large amounts of data like financial and healthcare industries, which actually need explainable ML the most. 
This is true for both local and global explanations. 
For local explanations, providing low latency implementation of explainable ML technique to the end-users has been challenging, and most organizations are unable to adopt it~\cite{bhatt2020explainable}. 
Several popular global explanation methods require getting local explanations for all datapoints and then either use an aggregation technique~\cite{global-aggregation,aggregation-tree,bhatt-aggregating} or select the most ``representative'' datapoints to provide a global picture~\cite{lime}. 
Running instance-wise local explainability methods for large datasets is time consuming and computationally expensive, and therefore not scalable. 
The global explanations generated using local explanations are also highly dependent on the aggregation method and therefore not reliable~\cite{global-aggregation,bhatt-aggregating}.


\begin{challenge}\label{ch:benchmarking}
Lack of standardized benchmarks
\end{challenge}

Fields like computer vision and natural language processing took off in strides after standard benchmarks were available to the community for specific problems, e.g. ImageNet for object recognition~\cite{deng2009imagenet}, SQuAD dataset for question answering~\cite{rajpurkar2016squad}. 

This paved the way for future works to be evaluated using these benchmarks and provided clear comparisons and improvements. 
Explainable ML lacks this standardized benchmarking. 
There do exist some attempts at creating these benchmarks, e.g. FICO Explainability challenge~\cite{fico}, ERASER~\cite{eraser-nlp-xai}, human-labeled benchmarks~\cite{mohseni2020humangrounded}, Medical Information Mart for Intensive Care~\cite{meng2021mimicif}. 
However, the need for deep domain knowledge to make sense of explanations due to the absence of clear evaluation metrics and adoptability remain a big hurdle in the progress of this endeavour. 



\begin{challenge}\label{ch:action}
Lack of research in the actionability from explanations. 
\end{challenge}

Most of current explainable ML research answers only one question: ``why was a particular prediction produced?'' There is lack of research in answering questions that arise after a stakeholder has received explanations for a prediction. 
For example, 1) if a box is being detected as a car, and the explanation highlights different sides of the box, what changes are required in the model to rectify this?, 2) if the most important feature for detecting a truck is the driver, how should the model be enforced to ignore it and focus on more general aspects, 3) if a person has good credit history and is yet being labeled into high-risk group, the explanation being their high credit score, how can we inform the model to abide by the intuitive rule of low risk label for such individuals. 
In industry, explainability has been promised as a step towards confident decision making. However, current explanations provide no actionable suggestions for either the developer of the model or the end-user. 
The explanations generated by most current explainability techniques are not causal in nature, adding to the problem of not being actionable if an end-user such as a clinician, wants to intervene and change the features in order to change outcome~\cite{lack-casusal}. 
Counterfactual explanations~\cite{verma2020counterfactual} give actionable suggestions to the end-users, but the requirements of the modellers are still unaddressed. 
The output of current explainability techniques are static in nature. 
Ideally, the modeller should be able to interact with the explainability technique. They should get explanations along with suggestions for bringing about the required changes, and then take actions to refine the model. This process should iteratively continue until satisfaction.

\begin{challenge}\label{ch:diversity}
Lack of stakeholder participation
\end{challenge}

The application of explainability techniques in high stakes domains like healthcare, finance, and criminal justice necessitates the need for teams where end-users interact with the developers of the model. 
In heavily regulated industries like finance, explainability has become crucial due to government regulations and guidelines. 
This in turn implies that professionals like lawyers and auditors are also the end users of explainable ML.
Indeed, it is known that different users comprehend definitions of ML \emph{fairness} differently~\cite{Grgic20:Dimensions,Saha20:Measuring}---it would be useful to perform similar studies of comprehension of ML \emph{explainability} as well. 
Hence, it is crucial to involve such users to understand their requirements and incorporate their expertise in designing explainability systems. 
Lack of inter-disciplinary teams in this traditionally computer science and mathematics oriented field is hindering its development and adoption. 



\begin{challenge}\label{ch:legal}
Lack of incentives and enforcements
\end{challenge}

Industry is interested in explainable ML both due to regulatory and reputational concerns. 
In Europe, the General Data Protection Regulation (GDPR) has laid down legal bindings for industries using ML in decision-making. The motivation behind GDPR was to ensure protection of the privacy, and the guarantee to fair and transparent access of resources to the end-users. 
It has been widely discussed that GDPR guarantees explanation to users in case of any adverse outcome. However, \citet{gdpr-nxai} argue the contrary and state that GDPR lacks the precise language to ensure a right for explanation. 
In the US, for example, DARPA has published a report with guiding principles for explainable ML~\cite{gunning2019darpa}; however, they are not legally binding.
Due to such ambiguous language and the laxity of enforcement, industries have little incentive to push for wide and effective adoption of explainable ML across their systems. 
This is further exacerbated by lack of resources in small and mid-sized firms. 
Thus much impetus for explainable ML comes from reputational concern for industries, especially in the US, rather than legal pressure. 
This must be addressed to ensure the required incentives for adoption and development of explainable ML.


\bibliographystyle{ACM-Reference-Format}
\bibliography{references}
\end{document}